\documentclass[conference]{IEEEtran}
\usepackage{multirow}

\usepackage{cite}

\ifCLASSINFOpdf
   \usepackage[pdftex]{graphicx}
   \DeclareGraphicsExtensions{.pdf,.jpeg,.png}
\else
\fi
\usepackage[cmex10]{amsmath}
\usepackage{array}

\usepackage[tight,footnotesize]{subfigure}

\usepackage{fixltx2e}
\def \constzeroindent {0cm}

\newenvironment{mycustomindent}[1]
{\setlength{\parindent}{#1}}
{\setlength{\parindent}{\constzeroindent}}

\begin{document}
%
\title{Dynamic Hand Gesture Recognition for Wearable Devices with Low Complexity Recurrent Neural Networks}

\author{\IEEEauthorblockN{Sungho Shin and Wonyong Sung}
\IEEEauthorblockA{Department of Electrical and Computer Engineering\\
Seoul National University\\
1, Gwanak-ro, Gwanak-gu, Seoul 08826 Korea\\
Email: shshin@dsp.snu.ac.kr, wysung@snu.ac.kr}
}


%


\maketitle

\begin{abstract}
Gesture recognition is a very essential technology for many wearable devices. While previous algorithms are mostly based on statistical methods including the hidden Markov model, we develop two dynamic hand gesture recognition techniques using low complexity recurrent neural network (RNN) algorithms. One is based on video signal and employs a combined structure of a convolutional neural network (CNN) and an RNN. The other uses accelerometer data and only requires an RNN. Fixed-point optimization that quantizes most of the weights into two bits is conducted to optimize the amount of memory size for weight storage and reduce the power consumption in hardware and software based implementations. 
\end{abstract}
%


%
\IEEEpeerreviewmaketitle

\section{Introduction}
Recently, many wearable devices have been developed for diverse applications, such as smart watches, Google Glass, and smart bands.  Since most wearable devices do not equip keyboards or wide touch screens, it is very necessary to employ speech or gesture recognition technologies. Although speech recognition can be more versatile, the gesture recognition can also be conveniently used for issuing simple commands.
\\
\indent  
There are several studies and applications that use the hand gesture technology in wearable devices. For example, Google Glass was controlled by hands and feet in~\cite{lv2014hand}, smart watches were controlled using flexible force sensors in~\cite{morganti2012smart}, and SixthSense employed a camera and a projector for interaction with real world~\cite{mistry2009sixthsense}.
\\
\indent 
Generally gestures can be classified into static and dynamic ones. Static gestures are usually represented by the hand shapes, while dynamic gestures are described according to hand movements~\cite{hasan2014static}. Gesture recognition can be conducted using signal from a camera or a force sensor. The former needs video processing while the latter analyzes time-varying multi-channel sensor output signal. When a force sensor is used, only the dynamic gesture recognition can be conducted.
\\
\indent
Recently, several hand gesture recognition algorithm have been developed~\cite{kim2009canonical,nagi2011max,neverova2013multi,mccartney2015gesture}. \cite{kim2009canonical} adopted the correlation between two videos which is called tensor canonical correlation analysis (TCCA). A convolutional neural network (CNN) is used for vision-based static hand gesture recognition for human robot interaction (HRI)~\cite{nagi2011max}. A multimodal gesture detection and recognition is studied using depth video, articulated pose, and audio stream~\cite{neverova2013multi}. They applied a CNN, a hidden Markov model (HMM) based speech recognizer, and a bag of word (BoW) to extract the multimodal features. For data fusion and gesture classification, they employed an Elman RNN. In~\cite{mccartney2015gesture}, an infrared LED data from Leap Motion Controller was used, where a CNN was applied for feature extraction and an HMM was adopted for time series recognition.  
\\
\indent
Neural networks are employed to many recognition applications including object detection and speech recognition~\cite{hwang2016character}.  However, neural networks demand heavy computation and large memory. For example, a long-short term memory (LSTM) RNN with the unit size of 256 demands a total of approximately 2.1 million (M) weights~\cite{gers2003learning}. For this reason, some of current wearable device applications such as speech recognition operate using servers that employ graphics processing units (GPUs) or multi-core systems consuming quite large power. Thus, it is greatly needed for wearable devices to operate neural network algorithms with only small power. 
\\
\indent 
In this study, we have developed dynamic gesture recognition techniques using fixed-point recurrent neural networks that are suitable for hardware or embedded system based implementations and low-power operation. Two gesture recognition algorithms are implemented; one uses the video signal from a camera and the other utilizes a three-axis accelerometer. Since dynamic gesture recognition needs to analyze the hand-movements, we employ LSTM RNNs. Also, a CNN is attached in front of the RNN for video based hand gesture recognition.
\\
\indent  
The RNN is optimized to minimize the hardware complexity. In order to minimize the memory size for weight storage, we conduct retrain based fixed-point optimization and successfully reduce most of the word-length into 2 bits~\cite{hwang2014fixed,shin2015fixed}.
\\
\indent
This paper is organized as follows. In Section~\ref{sec:fixed-point}, the proposed hand gesture recognition models and quantization procedure are given.  Experimental results are provided in Section~\ref{sec:experiment} and concluding remarks follow in Section~\ref{sec:conclusion}.
\section{Fixed-point RNN optimization}
\label{sec:fixed-point}
We employ two different kinds of dynamic hand gesture dataset. One is based on the image sequence~\cite{kim2009canonical}, and the other is the 3-axis acceleration data~\cite{costante2014personalizing}. Also, a fixed-point optimization scheme for these algorithms is  explained.  
\begin{figure}[!t]
\centering
\includegraphics[width=3.5in]{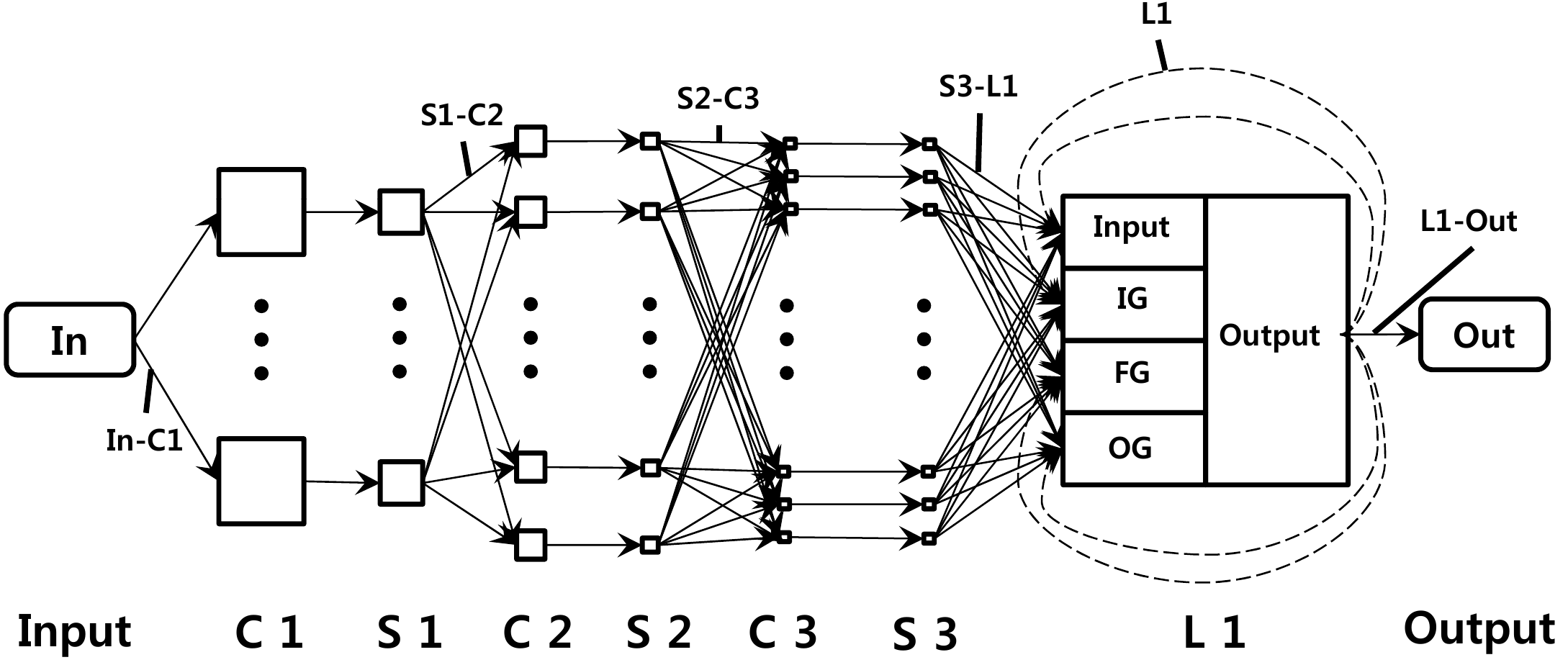}
 \caption{A structure of the three layer CNN and one LSTM RNN. The prefixes `C', `S' and `L' stand for convolution, subsampling and an LSTM layer, respectively. The prefixes `IG', `FG', `OG' represent input gate, forget gate and output gate of the LSTM layer. `IN-C1', `S1-C2', `S2-C3', `S3-L1', `L1', and `L1-Out' show weight groups for sensitivity analysis. Dotted lines and solid lines represent recurrent and forward paths, respectively.}
\label{fig_cnn_rnn}
\end{figure}
\subsection{Image Sequence Based Dynamic Hand Gesture Recognition}
For the image sequence based dynamic hand gesture recognition task, we employed a CNN-LSTM RNN structure. To generate hand shape features, three layer CNN architecture is chosen for its translational invariance properties. Our three layer CNN is similar to the one proposed by~\cite{cuda_convnet}. The specific network structure is depicted in~\figurename~\ref{fig_cnn_rnn}, which shows three convolution and pooling layers followed by an RNN layer. The input layer consists of 3072 (=3$\times$32$\times$32) linear units for handling the 32 by 32 input images with RGB channels. The first and the second convolution layers have 32 feature maps and the third convolution layer has 64 feature maps. These three layers have the same convolution kernel size, which is $5\times5$. The three pooling layers employ $2\times2$ overlapping max pooling. Thus, the CNN demands 79.2 kilo weights. Rectified linear units are adopted as for the activation functions. We employ the RNN layer to analyze gesture's temporal relation. The LSTM RNN can remember quite long past information in the sequence. As a result, HMM networks are not needed in this recognition model. The output layer consists of 9 softmax units which correspond to 9 target gesture behaviors. The total number of weights for the LSTM RNN is approximately 99.456 kilo weights. Therefore, a total of 0.714 MB (79.2 K and 99.456 K weights for CNN and RNN, respectively) memory space is required for the network model in a 32-bit floating-point format.   

\subsection{Acceleration Data Sequence Based Dynamic Hand Gesture Recognition}
The acceleration data sequence based dynamic hand gesture recognition model also employs the LSTM RNN structure. The standard LSTM uses three gates which are called the input gate, forget gate and output gate that can access and modify the memory cells. The activation functions for these three gates are the logistic sigmoid, and the input and output layers of the LSTM employ the hyperbolic tangent activation functions. This algorithm applies the acceleration data directly to the RNN. Therefore, this application is much simpler than the image sequence based dynamic hand gesture recognition. The input layer contains 3 linear units to receive the 3-axis acceleration data. One LSTM hidden layer with the size of 128 is used, and the output layer consists of 8 softmax units which correspond to 8 target gesture movements. An LSTM layer with $N$ units demands a total of $4N^{2} + 4NM + 7N$ weights where $M$ is the previous layer size~\cite{gers2003learning}. Therefore, the total number of weights is approximately 69 K, and as a result 276 KB memory space is needed for the network model when a floating-point format is used.      

\subsection{Retrain-Based Weight and Signal Quantization}
The quantization effects of signals or weights depend on a signal flow graph, and the influence of quantization can be represented as the sensitivity~\cite{sung1995simulation}. The weights and signals in each layer are grouped and each group employs the same quantization step size $\Delta$. To optimize $\Delta$, we adopt L2 error minimization criteria as suggested in~\cite{hwang2014fixed, anwar2015fixed}.
\\
\indent
Based on the quantization step size $\Delta$, sensitivity analysis for weights and signals is conducted layerwisely. \figurename~\ref{fig_cnn_rnn} shows the weight and signal grouping results. In this figure, `In-C1' is the first weight group between the input layer and the first convolution layer, `S3-L1' is the fourth weight group between the last pooling layer and the LSTM layer, `C1' is the signal group of the first CNN layer, and `L1' is the signal group of the LSTM layer.
\\
\indent
Since direct quantization does not show good performance, retraining on the quantization domain is performed. The RNN version of the retraining algorithm is introduced in~\cite{shin2015fixed}. 
\\
\indent
In our target networks, we use three different types of activation functions (logistic sigmoid, hyperbolic tangent and rectified linear unit). The output range of the logistic sigmoid function is between 0 and 1, that of the hyperbolic tangent is -1 and 1, and that of the rectified linear unit is theoretically 0 and $\infty$. Thus, the output signals for the logistic sigmoid and hyperbolic tangent activation functions are quantized with a fixed size of $\Delta$. However, the output value of the rectified linear unit can be unbounded. Therefore, the quantization step size $\Delta$ needs to be calculated in a similar method with the weights quantizer. The output signals of the rectified linear units are saved all over the training set to compute the proper quantization step size $\Delta$ with L2 error minimization.
\section{Experimental Results}
\label{sec:experiment}
The proposed algorithms are evaluated using two datasets obtained from wearable devices. One is the image based hand gesture recognition dataset and the other is the acceleration dataset from a 3-axis accelerometer. Advanced training techniques such as early stopping, adaptive learning rate, and Nesterov momentum are employed~\cite{jacobs1988increased,sutskever2013importance}.
\begin{figure}[t]
\begin{minipage}{1.0\linewidth}
  \centering
  \centerline{\includegraphics[width=6.5cm]{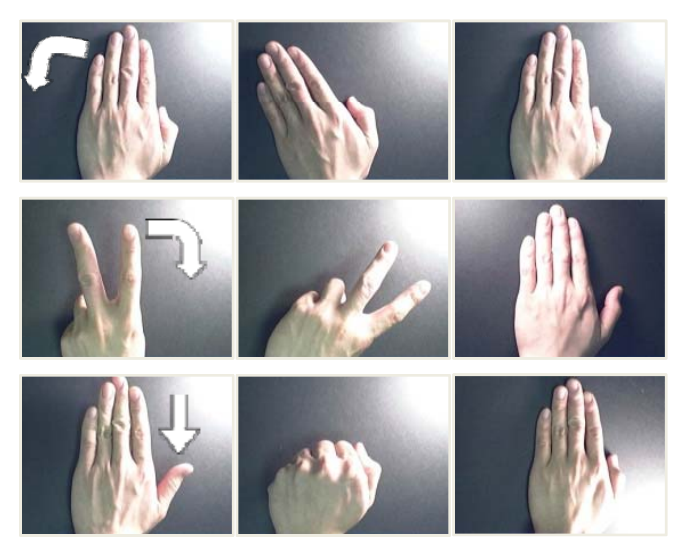}} 
  \centerline{(a) Cambridge Hand Dataset~\cite{kim2009canonical}}\medskip
 \end{minipage}
\begin{minipage}{1.0\linewidth}
  \centering
  \centerline{\includegraphics[width=6.5cm]{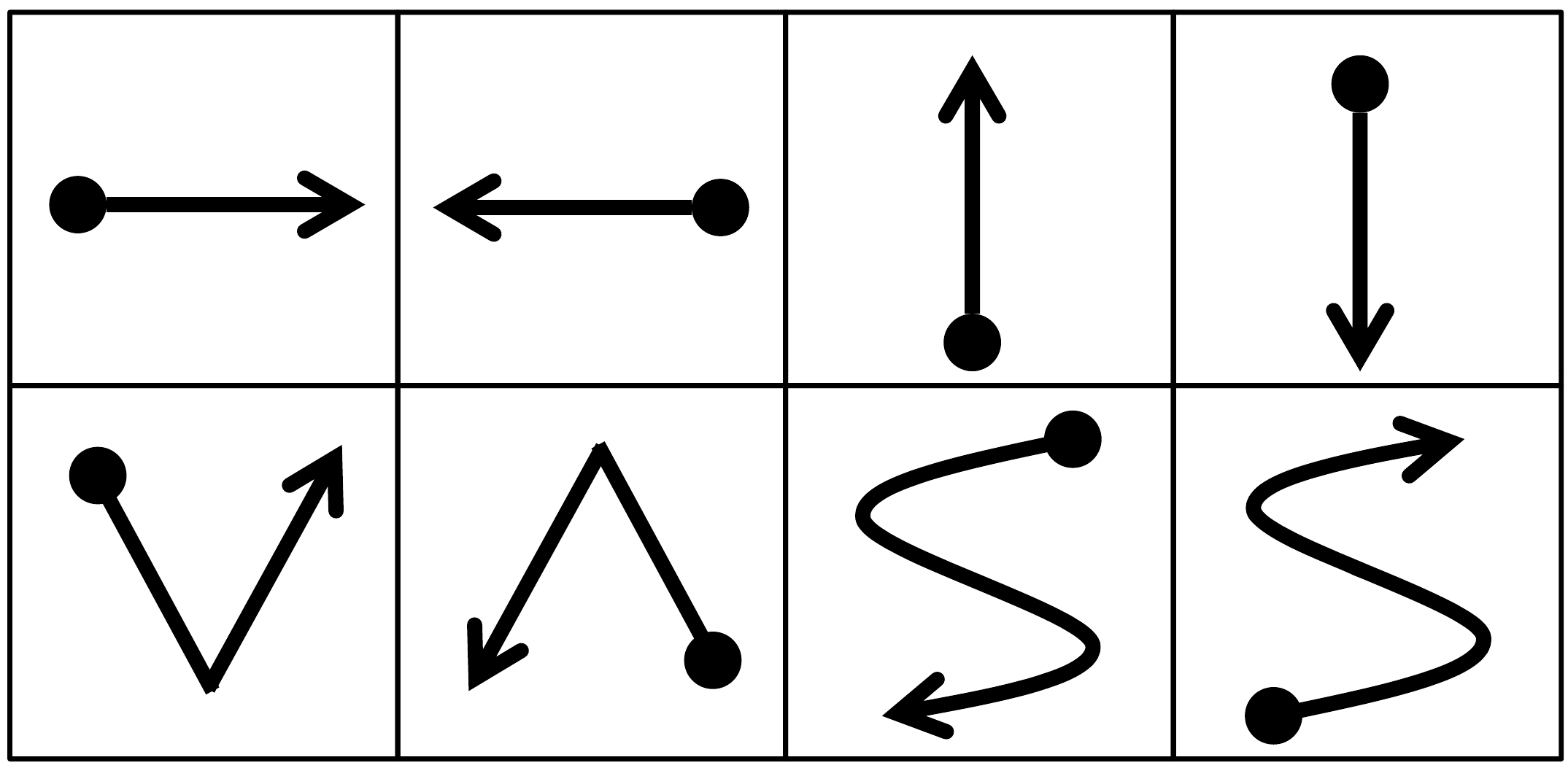}}
  \centerline{(b) SmartWatch Gestures Dataset~\cite{costante2014personalizing} }\medskip
 \end{minipage}
\caption{Dynamic hand gesture datasets for experiments; (a) is the image based dataset and (b) is the accelerometer based dataset.}
\label{fig:dataset}
\end{figure}
\subsection{Image Based Dynamic Hand Gesture Recognition}
Image based dynamic hand gesture recognition experiments were performed on the Cambridge-Gesture data base~\cite{kim2009canonical}. The data set consists of 900 image sequences of 9 gesture classes in QVGA (320 by 240), which are defined by 3 primitive hand shapes (flat, spread, and V-shape) and 3 primitive motions (left, right, and contract). Therefore, the target task for this data set is to classify different shapes as well as different motions simultaneously. Each class contains 100 image sequences (5 different illuminations $\times$ 10 arbitrary motions $\times$ 2 subjects). The dataset was divided into 60\% for the training (540 sequences), 20\% for the validation (180 sequences), and 20\% for the test (180 sequences) randomly. The ratios of the class labels are the same for the three sets.
\\
\indent
The network is trained using Fractal RNN library with training parameters that are 64 forward steps and 64 backward steps with 8 streams~\cite{hwang2015single}. Initial learning rate was $10^{-5}$ and the learning rate is decreased until $10^{-8}$ during the training. Momentum was 0.9 and AdaDelta was adopted for weights updating~\cite{zeiler2012adadelta}. We tried to find out the proper input size of the CNN with 64 by 64, 32 by 32, and 16 by 16 images. Their floating point classification error rates were 22.46\%, 22.79\%, and 73.70\% respectively. Therefore, the 32 by 32 image size is selected as the input image dimension. The network demands approximately 178.656 kilo weights. All experiments are repeated five times to consider their noise effects.
\begin{figure}[!t]
\centering
\includegraphics[width=3.5in]{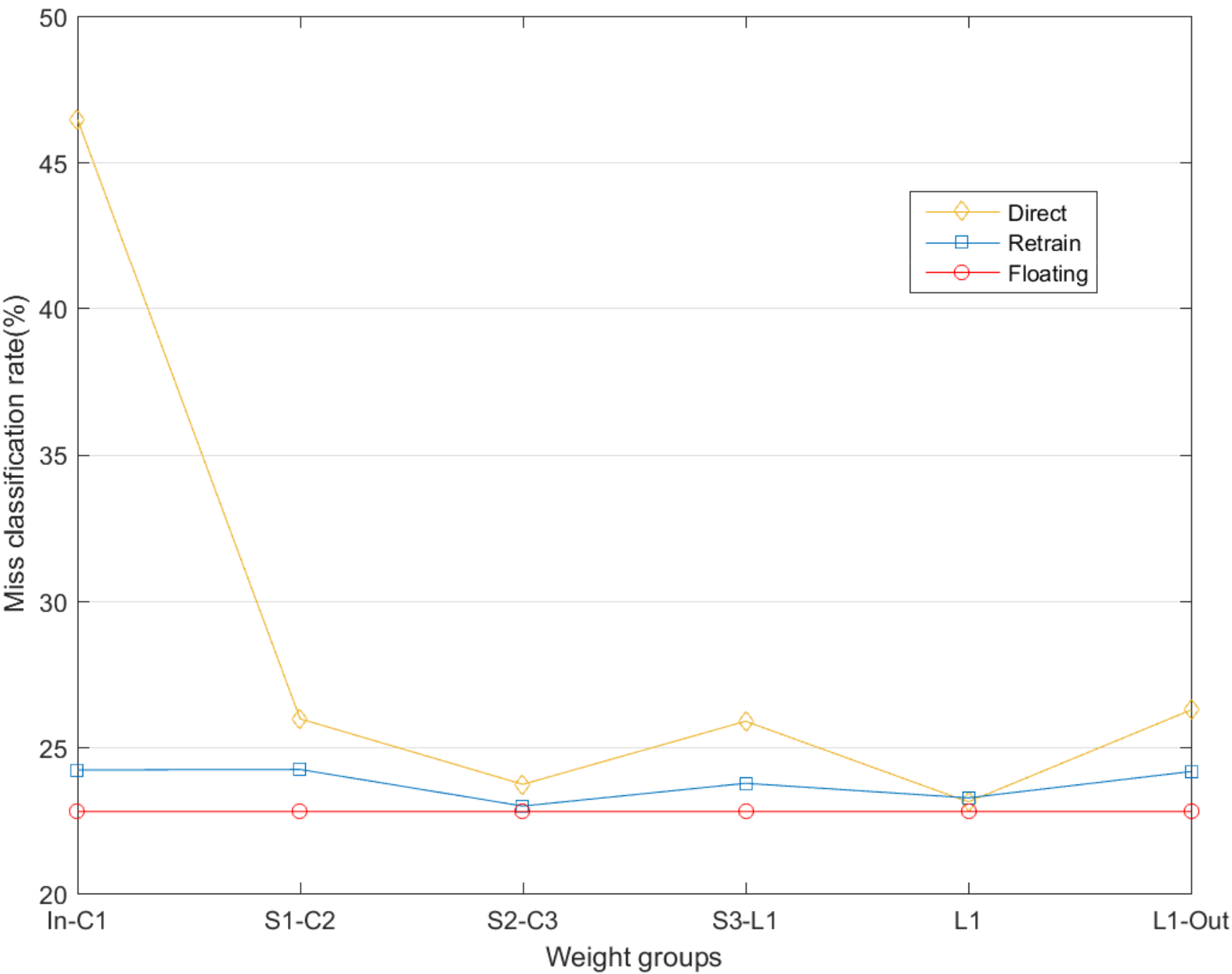}
 \caption{Layerwise weights sensitivity analysis results of the image based dynamic hand gesture recognition example. The red line indicates the floating-point results, the blue line represents the direct quantization result, and the green line shows the retraining results.}
\label{fig_weight_sense_image}
\end{figure}
\\
\indent 
\figurename~\ref{fig_weight_sense_image} shows the layerwise fixed-point sensitivity analysis for the weight groups. The original miss classification rate was 22.79\% for the test set. The results indicate that the most sensitive weight groups are `In-C1' and `L1-Out'. However, the final fixed-point network employs only 2 bits for all the weight groups because their sensitivity difference is small.
\begin{table}[!]
\renewcommand{\arraystretch}{1.2}
\centering
\caption{Layerwise sensitivity analysis results for signal groups in the image sequence based model. Each layers is quantized in two bits for sensitivity analysis. The numbers in the table represent the miss classification rate (\%) of the test set. `D' means direct quantization results and `R' represents the results after retraining.}
\label{table_layer_image}
\resizebox{\columnwidth}{!}{
\begin{tabular}{c|l|c c c c c c c c c}
\hline
\multicolumn{2}{c}{}  & In    & C1    & S1    & C2    & S2    & C3    & S3    & L1  & All  \\ \hline\hline
\multicolumn{2}{c}{D} & 25.97 & 28.41 & 27.25 & 31.80 & 29.61 & 23.49 & 23.68 & 23.86 & 58.24 \\ \hline
\multicolumn{2}{c}{R} & 24.31 & 20.15 & 24.26 & 23.49 & 22.56 & 23.86 & 22.90 & 23.91 & 24.40 \\ \hline
\end{tabular}}
\end{table}
\tablename~\ref{table_layer_image} depicts the results of the signal group sensitivity analysis. All the signal groups also employ only two bits after retraining. Note that some layers even show better performance when compared to the floating point results. We obtained a fully quantized network with these two sensitivity results. All the weight and signal groups were quantized by using only two bits, and the miss rate was 25.04\%. With this quantization, the memory space saved is 93.75\% (=30/32) when compared to a floating-point implementation. The total number of multiplications for each layer is 56.448 M (C1), 76.8 M (C2), 1.536 M (C3), 1.966 M (L1) and 34.56 K (Out), respectively, for real-time operation (30 Hz). The memory space needed is 44.625 KB. Therefore, our model can efficiently be implemented in embedded systems such as Cortex-A9 (128 KB - 8 MB L2 cache), since the whole weights memory can be stored in the on-chip L2 cache.
\subsection{Accelerometer Based Dynamic Hand Gesture Recognition}
An accelerometer based dynamic hand gesture recognition model was trained using the SmartWatch Gestures Dataset~\cite{costante2014personalizing}. The data set has been collected to evaluate several gesture recognition algorithms for interacting with mobile applications using arm gestures. Eight different users performed twenty repetition of twenty different gestures for a total of 3200 sequences. Each sequence contains acceleration data from the 3-axis accelerometer of a first generation Sony SmartWatch. Original dataset contains 20 motions, but eight motions which are depicted in \figurename~\ref{fig:dataset} (b) are enough as a wearable device controller. The training, validation, and test sets are divided randomly into 50\%, 20\%, and 30\% respectively.
\\
\indent 
The RNN training method is the same with~\cite{hwang2015single}. Initial learning rate was $10^{-5}$ and it is decreased until $10^{-7}$ during the training procedure. Momentum was 0.9 and AdaDelta was employed for weight updating. We tried to find out the proper network size of the LSTM layer. The network sizes of 32, 64, 128 and 256 were considered. Since the entire number of the dataset is too small to training the LSTM RNN, the test set noise was very large. Therefore we conduct the experiments 10 times for each network size. The floating point training results were 36.02$\pm24.40$\% (mean$\pm$standard deviation), 22.26$\pm4.00$\%, 18.29$\pm4.86$\%, and 16.68$\pm5.72$\% for the 32, 64, 128 and 256 network size, respectively. We chose the LSTM layer with the size of 128, and the floating-point weights that show best error rate on the test set, which was 11.43\%. 
\begin{table}[t]
\renewcommand{\arraystretch}{1.2}
\centering
\caption{Layerwise sensitivity analysis results in the accelerometer based model. The numbers in the table are miss classification rates (\%) of the test set. All groups are quantized in two bits except `L1' signal group. `L1(2)' and `L1(3)' represent that the signal sensitivity analysis was performed in two and three bits, respectively.}
\label{table_acc}
\begin{tabular}{c c c c }
\hline\hline
\textbf{WEIGHT}  & \textbf{In-L1} & \textbf{L1}    & \textbf{L1-Out} \\ \hline
Direct quantization & 10.99 & 11.39 & 10.77  \\ \hline
Retrain based & 11.11 & 11.31 & 11.56  \\ \hline\hline
\textbf{SIGNAL}  & \textbf{In}    & \textbf{L1(2)} & \textbf{L1(3)}  \\ \hline
Direct  & 12.31 & 88.84 & 88.69  \\ \hline
Retrain & 11.27 & 33.64 & 12.56  \\ \hline
\end{tabular}
\end{table}
\\
\indent
\tablename~\ref{table_acc} shows the layerwise sensitivity analysis results for weights and signals. The floating points classification error rate was 11.43\%. Since the training set is too small for training the LSTM RNN, overfitting was observed when retraining the network in the quantization domain. The signal sensitivity analysis result shows that the `L1' layer group does not yield good performance with only two bits quantization levels. It needs three or more bits to obtain acceptable results.
\\
\indent
We next try fixed-point optimization of all signals and weights using the sensitivity analysis results. We applied two bits to all weight and signal groups and its miss classification rate was 32.77\%. Therefore, we assigned more bits to the `L1' signal group and obtained 28.69\% in three bits and 11.43\% with four bits. The whole memory space can be reduced by 93.75\% when compared to a floating point implementation, since the memory space for weights is only 17.25 KB. The total number of multiplications was 690 K for real time operation (10~Hz). Our model can be efficiently implemented in embedded systems, since the whole weights can be stored  in the cache memory.

\section{Concluding Remarks}
\label{sec:conclusion}
This work investigates the fixed-point implementation of LSTM RNNs for dynamic hand gesture recognition using two different datasets. The RNN based implementations show good results and can be improved further by using more training data. The retrain-based fixed-point optimization greatly reduces the word length of the weights and signals. By this optimization, the required memory space for weights can be reduced to only 6.25\% compared to floating-point implementations.  The optimized fixed-point network can show much better energy performance in embedded implementations because the reduced memory size enables purely on-chip memory based operations.


\section*{Acknowledgment}
This work was supported in part by the Brain Korea 21 Plus Project and the National Research Foundation of Korea (NRF) grant funded by the Korea government  (MSIP) (No. 2015R1A2A1A10056051).



\bibliographystyle{IEEEtran}
%

\bibliography{refs}


\end{document}